\documentclass[10pt,twocolumn,letterpaper]{article}

\usepackage[pagenumbers]{cvpr} % To force page numbers, e.g. for an arXiv version

\usepackage{graphicx}
\usepackage{amsmath}
\usepackage{amssymb}
\usepackage{booktabs}
\usepackage{color}
\usepackage{rotating}
\usepackage{makecell}

\definecolor{cardinal}{rgb}{0.77, 0.12, 0.23}
\definecolor{officegreen}{rgb}{0.0, 0.5, 0.0}
	\definecolor{lightbrown}{rgb}{0.71, 0.4, 0.11}

\usepackage[pagebackref,breaklinks,colorlinks]{hyperref}

\usepackage[capitalize]{cleveref}
\crefname{section}{Sec.}{Secs.}
\Crefname{section}{Section}{Sections}
\Crefname{table}{Table}{Tables}
\crefname{table}{Tab.}{Tabs.}

\def\cvprPaperID{8} % *** Enter the CVPR Paper ID here
\def\confName{CVPR}
\def\confYear{2023}

\begin{document}

%%%%%%%%% TITLE - PLEASE UPDATE
%Toward Super-Resolution for Appearance-Based Gaze Estimation
\title{Toward Super-Resolution for Appearance-Based Gaze Estimation}

\author{Galen O'Shea\\
Carleton Unviersity\\
Ottawa, ON, Canada \\
{\tt\small galenoshea@cmail.carleton.ca}
% For a paper whose authors are all at the same institution,
% omit the following lines up until the closing ``}''.
% Additional authors and addresses can be added with ``\and'',
% just like the second author.
% To save space, use either the email address or home page, not both
\and
Majid Komeili\\
Carleton Unviersity\\
Ottawa, ON, Canada \\
{\tt\small majid.komeili@carleton.ca}
}
\maketitle

%%%%%%%%% ABSTRACT
\begin{abstract}
   Gaze tracking is a valuable tool with a broad range of applications in various fields, including medicine, psychology, virtual reality, marketing, and safety. Therefore, it is essential to have gaze tracking software that is cost-efficient and high-performing. Accurately predicting gaze remains a difficult task, particularly in real-world situations where images are affected by motion blur, video compression, and noise. Super-resolution has been shown to improve image quality from a visual perspective. This work examines the usefulness of super-resolution for improving appearance-based gaze tracking. We show that not all SR models preserve the gaze direction. We propose a two-step framework based on SwinIR super-resolution model. The proposed method consistently outperforms the state-of-the-art, particularly in scenarios involving low-resolution or degraded images. Furthermore, we examine the use of super-resolution through the lens of self-supervised learning for gaze prediction. Self-supervised learning aims to learn from unlabelled data to reduce the amount of required labeled data for downstream tasks. We propose a novel architecture called ``SuperVision'' by fusing an SR backbone network to a ResNet18 (with some skip connections). The proposed SuperVision method uses 5x less labeled data and yet outperforms, by 15\%, the state-of-the-art method of GazeTR which uses 100\% of training data. We will make our code publicly available upon publication.
   
\end{abstract}

%%%%%%%%% BODY TEXT
\section{Introduction}

Gaze estimation has been regarded as an valuable research area with a broad range of applications. Research into gaze tracking has often been divided into model-based and appearance-based approaches. While model-based methods rely on specialized hardware, appearance-based methods only require a camera \cite{hansen2009eye}. Although there have been improvements in appearance-based gaze tracking, accurately predicting gaze remains a difficult task, particularly outside controlled environments and curated datasets \cite{alberto2014geometric, xiong2014eye, sun2015real, funes2014eyediap}. Furthermore, research in appearance-based gaze tracking compensate for the lack of specialized hardware by using large and complex machine learning algorithms to extract information \cite{zhang2015appearance, ali2020deep, bao2021adaptive}, leaving the data overlooked.

\begin{figure}[t]
\centering
\includegraphics[width=0.45\textwidth]{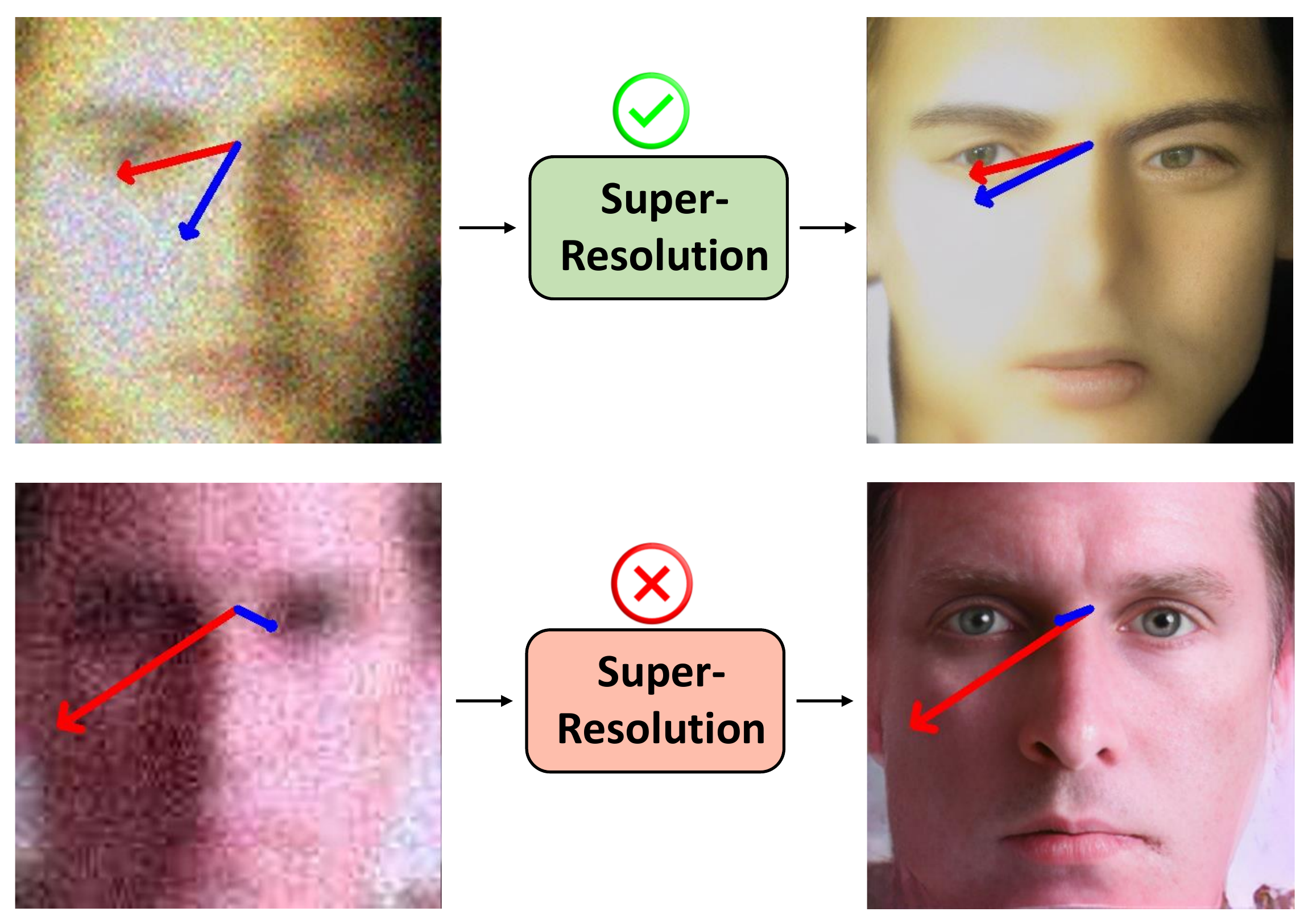}
\caption{Top: an example of super-resolution improving gaze prediction. Bottom: an example of super-resolution degrading the gaze prediction by hallucinating a frontal gaze. The blue line is the gaze prediction and and the red line is the ground truth. Super-resolution is done using SwinIR \cite{liang2021swinir} and GFP-GAN \cite{wang2021towards} for the top and the bottom rows respectively. Gaze Estimation is done using Full-Face \cite{zhang2017s}.}
\label{SwinIR}
\end{figure}

Super-resolution is a technique used to increase the resolution of an image beyond the original image resolution. Super-resolution has been shown to improve image quality from a visual perspective \cite{ledig2017photo, wang2018esrgan, wang2021real, wang2021towards, liang2021swinir, saharia2021image}. For instance, SR can increase detail and clarity while removing degradations such as motion blur, file compression, and pixelation. Super-resolution has been shown to be useful in many applications, including medical imaging \cite{shi2013cardiac}, remote sensing \cite{ling2019super}, surveillance and video processing \cite{liu2022video}. While SR has been shown to create visually appealing images, its usefulness for gaze prediction has not been investigated. Figure \ref{SwinIR} shows SR as a preprocessing step for gaze prediction. We rigorously examine the usefulness of SR for gaze prediction.

Furthermore, we examine the use of SR through the lens of self-supervised learning for gaze prediction. Self-supervised learning allows a machine learning model to learn from unlabelled data which can be easier and less expensive to acquire than labeled data. For example, a self-supervised model might be trained to predict the rotation or color of an image which can be done without the need for manual annotation or labelling. Once the self-supervised model is trained, it can be fine-tuned for a specific task using a smaller amount of labeled data. Gaze prediction, like many traditional supervised learning tasks, requires a large amount of labeled data. A sample efficient model opens up possibility of building gaze models for scenarios where obtaining a large labelled gaze dataset is challenging, for example gaze tracking in infants \cite{franchak2016free}, older adults \cite{chapman2006evidence} or various types of animals. Gaze tracking in animals can be useful for studying their behaviour, cognition, and perception \cite{wiltschko2015mapping, ogura2020dogs, guo2003monkeys}. 

Contributions of this paper are as follows. 
\begin{enumerate}
    \item  We rigorously examine the usefulness of SR when used as a preprocessing step for appearance-based gaze prediction. We show that not all SR models preserve the gaze direction.
    \item We propose a two-step framework based on SR and achieve state-of-the-art results on the MPIIFaceGaze \cite{zhang2015appearance} dataset. Additionally, the performance of the proposed method was evaluated on low-resolution images and various degradations, and the proposed method consistently outperformed other methods.
    \item We propose a novel architecture called “SuperVision” by fusing an SR backbone network to a ResNet (with some skip connections). The proposed SuperVision method uses only 20\% of the labeled data and yet outperforms, by 15\%, the state-of-the-art method of GazeTR \cite{cheng2021appearance} which uses 100\% of data.
\end{enumerate}

\section{Related Work}
\subsection{Traditional Approach}
In the field of gaze tracking research, there are two primary approaches: model-based and appearance-based methods \cite{hansen2009eye}. Model-based methods aim to predict gaze by using either a geometric 3D eye model that analyzes the corneal reflection \cite{nakazawa2012point}, or shape-based methods that utilize the pupil centre \cite{valenti2011combining} or contours of the iris \cite{funes2014eyediap}. Advancements in corneal-reflection techniques have enabled the transition from fixed head positions \cite{morimoto2002detecting} to multiple head poses and lighting conditions \cite{zhu2006nonlinear}. However, due to the complexity of these methods, researchers often employ costly and specialized hardware (e.g., depth sensors) that may not be practical for general-purpose gaze tracking \cite{alberto2014geometric, xiong2014eye, sun2015real, funes2014eyediap}. Additionally, although these methods exhibit outstanding performance in a controlled laboratory setting, they are less reliable in low light conditions and unconstrained environments.

Appearance-based methods frame gaze estimation as a regression problem by mapping gaze images to a corresponding gaze vector, while avoiding the need for specialized hardware and using only a camera. However, early appearance-based methods required time-consuming head-pose calibration for each participant, which led to research on reducing the number of training examples using semi-supervised Gaussian regression methods \cite{williams2006sparse} and finding an optimal set of training samples using adaptive linear regression \cite{lu2014adaptive}. Despite these efforts, calibration remained insufficient as models failed to generalize to new subjects and head poses, leading to research on solving head pose and subject-related issues using a pose-based clustering method \cite{sugano2008incremental} and compensating for bias via regression \cite{lu2014learning}. To address free head motion, eye image synthesis was later employed \cite{lu2015gaze}. Generalization problems were further handled using cross-subject training methods \cite{mora2013person} and learning-by-synthesis methods \cite{sugano2014learning}.

\subsection{Deep Learning Approach}
Early approaches to gaze tracking faced significant challenges in adapting to new subjects and positions despite efforts to enhance their performance. As a result, research shifted to deep learning approaches such as the Multimodal Convolutional Neural Network (CNN), which concatenate eye images with a head pose estimate \cite{zhang2015appearance}. In subsequent work, the authors designed a VGG-inspired architecture that extended their previous work \cite{zhang2017mpiigaze}. Other methods included dilated-convolutions to extract high-level features without reducing spatial resolution \cite{chen2018appearance}. In related studies on multi-stream CNNs, researchers adopted data fusion, which involved merging datasets while maintaining separate validation sets to evaluate the performance of each dataset independently \cite{ali2020deep}. Although state-of-the-art performance was achieved, the variation in accuracy across datasets was attributed to differences in the resolution of the eye patch. The researchers suggested that deep learning models for gaze estimation could benefit from higher accuracy if the datasets had more pixels within each eye patch \cite{ali2020deep}.

\subsubsection{Attention Approach}
Studies have indicated that combining eye feature maps can enhance the accuracy of gaze tracking. For instance, a study utilized a VGG-16 network per eye, with concatenated downstream features \cite{Fischer_2018_ECCV}. Other works, such as \cite{chen2018appearance}, employed quad-stream architecture to extract singular and joint features from both eyes. Subsequently, researchers explored the use of the attention mechanism to extract joint eye features. In one novel approach, \cite{bao2021adaptive} proposed an Adaptive Feature Fusion technique that stacked eye feature maps based on their similarity, using a self-attention mechanism. More recently, transformer architectures have emerged as dominant in the field, including for gaze tracking. The current state-of-the-art model, GazeTR, uses a hybrid CNN and transformer architecture \cite{cheng2021gaze}.

\subsubsection{Generative Adversarial Networks Approach}
In the past, the predominant approach to solving the gaze estimation problem has been to develop complex models, with limited attention given to improving the quality of the data. While a few studies have utilized generative adversarial networks (GANs) to enhance data quality by improving lighting conditions \cite{kim2020gaze} or removing artifacts from glasses \cite{rangesh2020driver}, these niche approaches have demonstrated limited practical success, as they do not address the fundamental challenge of obtaining high-quality datasets.

\subsection{Super-Resolution}
The first application of GANs to super-resolution was in SRGAN, which outperformed prior methods and achieved state-of-the-art results \cite{ledig2017photo}. The authors attributed this, in part, to their use of a perceptual loss function that accounted for perceptual similarity instead of just similarity in pixel space. At that time, a common problem with super-resolution was the presence of artifacts when upsampling. ESRGAN addressed this issue by identifying that batch normalization layers tended to create unwanted artifacts \cite{wang2018esrgan}. They also improved the perceptual loss function and used Residual-in-Residual Dense Blocks to generate more realistic images consistently. Additionally, they later proposed REAL-ESRGAN, which incorporated a u-net discriminator and spectral normalization \cite{wang2021real}. They also developed a complex degradation method that used only synthetic data for Real-World Degradations.

Blind face restoration is often challenging because it requires prior knowledge, such as facial geometry, to restore realistic details. GFP-GAN utilized generative facial prior in the image restoration process and achieved realistic details and state-of-the-art results \cite{wang2021towards}. Other attempts at image restoration have employed transformers. For example, one study utilized a shifted window transformer as a deep feature extractor in a model called SwinIR, which achieved state-of-the-art results with up to 67\% fewer parameters \cite{liang2021swinir}. However, GANs for super-resolution are becoming less popular, as the current state of the art is based on an iterative refinement method \cite{saharia2021image}, which significantly outperforms previous works.

While most existing methods typically employ classical degradation methods such as downsampling to generate low-resolution images, Real-ESRGAN utilized synthetic data and a complex degradation model that aimed to simulate real-world complex degradations \cite{wang2021real}. Their degradation model involved combining multiple classical degradations, including Gaussian filters for blurring, downsampling using interpolating methods, adding Gaussian, colour, and other types of noise, and reducing quality through JPEG compression. However, other researchers found that the degradation model used by Real-ESRGAN lacked diversity, and they addressed this issue by expanding the model through random shuffling of the process, adding different levels of noise and compression, and introducing processed camera sensor noise and RAW image noise \cite{zhang2021designing}.

\subsection{Self-Supervised Learning}
Self-supervised learning has emerged as a powerful technique for learning rich and meaningful representations from unlabeled data. This technique involves training a model to predict a pretext task from the input data, which then results in learning useful representations that can be transferred to downstream tasks. One popular form of self-supervised learning is contrastive learning, which aims to learn representations by contrasting positive and negative pairs of samples \cite{DBLP:journals/corr/abs-1911-05722, DBLP:journals/corr/abs-2006-07733, DBLP:journals/corr/abs-2005-10243}. SimCLR \cite{pmlr-v119-chen20j} has been one of the most successful and extensively used approaches in this field. It has achieved state-of-the-art results on various benchmarks, including ImageNet and COCO. The authors have demonstrated that this technique is effective in pretraining models for several downstream tasks, such as object detection, instance segmentation, and semantic segmentation.

%MK: TEMPRARY 
%\subsection{Blind Super-Resolution Preprocessing}
%Prior studies have proposed that using a dataset with higher pixel density in the eye regions could enhance the accuracy of gaze estimation using deep learning \cite{ali2020deep}. To achieve this, the propose method employs a SR as a preprocessing step to enhance the quality and pixel density. Although several SR models are available, our research focuses on GFP-GAN \cite{wang2021towards} and SwinIR \cite{liang2021swinir} due to their high performance in PSNR and SSIM, which are commonly used to evaluate the quality of generative image techniques. These models adopt distinct approaches to SR; GFP-GAN leverages facial priors for image restoration, while SwinIR utilizes a transformer. Additionally, we evaluated two SR scales, namely 2x and 4x resolution, for upscaling images. This decision was partly due to hardware constraints when 4x upscaling from size $224 \times 224$.

\section{On the Usefulness of SR for Gaze Prediction}

\subsection{Not All SR Models Preserve Gaze}
Prior studies have suggested that using face images with higher pixel density in the eye regions could enhance the accuracy of gaze estimation using deep learning \cite{ali2020deep}. Super-resolution refers to the process of increasing the resolution of an image by recovering or generating high-resolution images from low-resolution inputs. It has been shown to be useful in various computer vision tasks, but its usefulness for gaze prediction has not been studied. In particular, while SR can increase detail and clarity, it is not apparent if the SR process alters the gaze in the resulting image. Along this line, we examine two different SR models.

Although several SR models are available, we focused on GFP-GAN \cite{wang2021towards} and SwinIR \cite{liang2021swinir} due to their high performance in PSNR and SSIM, which are commonly used to evaluate the quality of generative image techniques. These models adopt distinct approaches to SR; GFP-GAN leverages facial priors for image restoration, while SwinIR utilizes a transformer. Furthermore, GFP-GAN was pretrained using the FFHQ dataset \cite{karras2019style} while SwinIR was pretrained on DIV2K \cite{Ignatov_2018_ECCV_Workshops} and Flickr2K \cite{timofte2017ntire}.
For gaze, we use the Full-Face model \cite{zhang2017s} which is widely considered as the basis for much of the current research in appearance-based gaze tracking. We use the MPIIFaceGaze dataset \cite{zhang2015appearance} which comprises 45,000 images of size $448 \times 448$, gathered from 15 subjects of different ethnicities under natural lighting conditions. To evaluate the efficacy of the proposed method, the images are first downsampled to a size of $112 \times 112$ allowing room for 4x upsampling. To replicate real-world degradations, we used the method used in \cite{zhang2021designing} to degrade the images. Subsequently, we applied one of the pretrained SR models (GFP-GAN \cite{wang2021towards} or SwinIR \cite{liang2021swinir}) to upscale the images back to size $448 \times 448$. To ensure that upscaling did not serve as a confounding variable, we also upscaled the dataset using bicubic interpolation to create a baseline. We then train and evaluate Full-Face with each of the three new datasets using a leave-one-out cross-validation approach. This involved training on 14 of the 15 participants and testing on the one participant that was excluded. We repeated this process for each participant to calculate the average point of gaze (POG) error. The epoch with the highest test performance in Table \ref{exp1_tab}.

The aim of the initial experiment was to assess how various SR techniques compare to an interpolation baseline when applied to images with and without degradations. The experiment utilized Full-Face \cite{zhang2017s} as the gaze model and downsampled data of size 112. 

The results of the scenario without degradations, shown in Table \ref{exp1_tab} were unexpected, with GFPGAN performing 8.4\% worse than the baseline, while SwinIR-4x exhibited a 2.1\% improvement. In the degraded scenario, all interpolation methods achieved mediocre results compared to the first scenario, likely due to the presence of complex degradations. Once again, GFP-GAN under-performed relative to the baseline, however, SwinIR-4x outperformed the baseline by 7.3\%. The results indicate that while GFP-GAN can generate visually sharp images, it is inadequate for the task of gaze estimation. On the other hand, SwinIR outperformed the baseline with both regular and degraded images. The most interesting finding is that SwinIR achieved a 5.2\% improvement in relative performance indicating that it was more effective in degraded scenarios, emphasizing the applicability of SR in real-world situations. While this experiment shows a large contrast in SR methods, it has also proved that SR can be an effective tool for gaze estimation.

\begin{table}[h]
\centering
\begin{tabular}{ccc}
\hline
Degradation Type                      & Interpolation Type    &  POG \\ \hline
\multirow{3}{*}{None}                 & Interpolation & 4.20$^{\circ}$ \\
                                      & GFP-GAN-4x            & 4.59$^{\circ}$ \\
                                      & SwinIR-4x             & \textbf{4.11$^{\circ}$} \\ \hline
\multirow{3}{*}{Complex Degradations} & Interpolation & 5.47$^{\circ}$  \\
                                      & GFP-GAN-4x            & 5.76$^{\circ}$ \\
                                      & SwinIR-4x             & \textbf{5.10$^{\circ}$} \\ \hline
\end{tabular}
\caption[SR on Full-Face]{Comparison of different SR methods for gaze prediction. GFP-GAN performs worse than a simple interpolation. SwinIR significantly improves the gaze prediction. In all cases, Full-Face is used as the gaze model.}
\label{exp1_tab}
\end{table}

\begin{figure}[h]
\centering
\includegraphics[width=0.45\textwidth]{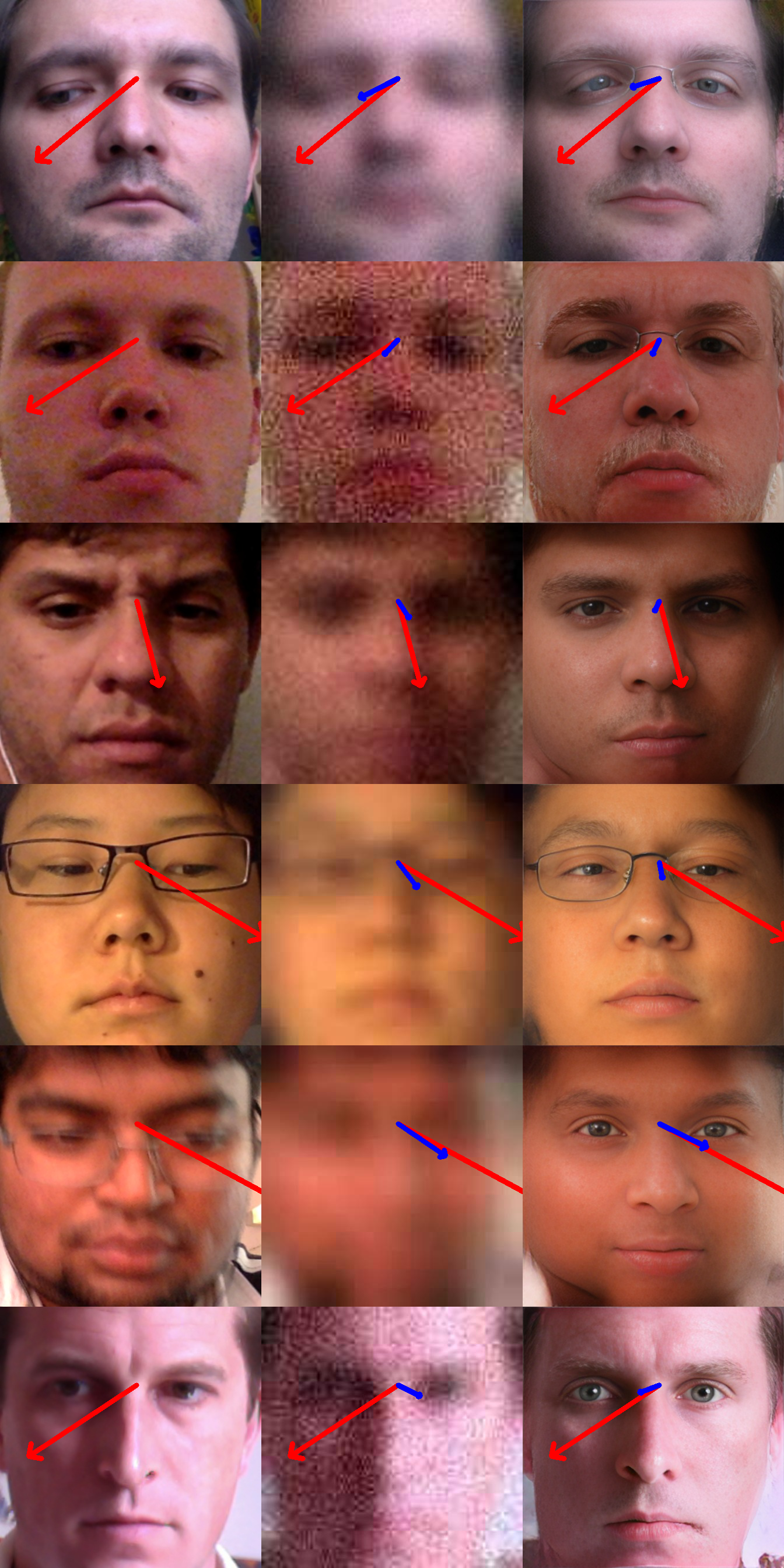}
\caption[Mode Collapse]{ Gaze Mode Collapse. Left: original images. Centre: degraded images. Right: restored images using GFP-GAN. GFP-GAN tends to generate images where subject starring at the camera. The red arrow depicts the ground truth gaze vector while the blue is predicted.}
\label{mode_collapse_fig}
\end{figure}

%\begin{figure}[h]
%\centering
%\includegraphics[width=0.45\textwidth]{images/haluc.png}
%\caption[In the Face of Uncertainty]{ GFP-GAN hallucinations. The left column shows the original image, the middle column degrades the left image, and GFP-GAN %restores the middle image in the last column. The red line shows the ground truth gaze vector while the blue is the gaze prediction.}
%\label{haluc_fig}
%\end{figure}
As demonstrated in Table \ref{exp1_tab}, not all SR methods are suitable for task dependent image restoration. GFP-GAN's poor gaze tracking performance might be attributed to the use of facial priors, which bias the model into hallucinating facial features from noise. Additionally, "in the face of uncertainty", GFP-GAN's dependence on facial priors resulted in restored images having a similar gaze focused on the centre of the screen. This gaze hallucination is most evident when the input images are significantly degraded, indicating that GFP-GAN is susceptible to mode collapse. The issue of mode collapse is apparent in the hallucinated gazes depicted in Figure \ref{mode_collapse_fig}. Unlike other restoration approaches, GFP-GAN was trained on the FFHQ dataset \cite{karras2019style}, which contains mostly images of people looking directly at the camera. Since GFP-GAN does not attempt to maintain the original gaze and is influenced by facial priors, the reconstructed faces have their gazes centred and therefore "looks can deceiving".

\subsection{Improving Gaze Estimation with SR}

%The goal of the second experiment was to assess the efficacy of SR when the gaze model already achieves SOTA performance. Furthermore, the experiment was designed to compare the effectiveness of SR when the upscaled image is downscaled back to its original size to determine if additional information is lost through compression. 
 
In this section, to further improve the gaze prediction results, Full-Face is replaced with the state-of-the-art model, GazeTR \cite{cheng2021gaze} which uses a hybrid CNN and transformer architecture. Considering the poor performance of GFP-GAN, as discussed in the previous section, only SwinIR will be used for the remaining experiments. We demonstrate that the resulting framework surpasses GazeTR and produces state-of-the-art results on the MPIIFaceGaze dataset \cite{zhang2015appearance}. To reproduce claims by the authors, GazeTR was pretrained on ETH-XGaze which is a dataset consisting of over one million high-resolution gaze images presented in multiple head poses. The dataset was collected from 110 subjects using multiple SLR cameras, lighting conditions and a calibrated ground truth \cite{zhang2020eth}. 

 The proposed method (second section in Table \ref{exp2_tab}) uses additional preprocessing techniques.``SwinIR-2x Downsampled" upscales the initial $224 \times 224$ data (Table \ref{exp2_tab} ``Input" column) by 2x to a size of $448 \times 448$ followed by downsampling back to $224 \times 224$ for a fair comparison to GazeTR. The $224 \times 224$ data is also upsampled to $448 \times 448$ using bicubic interpolation to give a fair comparison. "SwinIR-2x" uses SR to upscale the data 2x, and is kept at size $448 \times 448$ to compare to the interpolated baseline mentioned previously. Finally, the original $448 \times 448$ high-resolution data is used with GazeTR as an upper limit on performance.

The first section in Table \ref{exp2_tab} demonstrates prior works performance on images of size $224 \times 224$, while the second section demonstrates results from our second experiment. The results show "SwinIR-2x Downsampled" achieved a POG of 3.94$^{\circ}$ and demonstrated a 1.5\% improvement in performance compared to GazeTR's 4.00$^{\circ}$. The bicubic interpolation $448 \times 448$ baseline achieved a POG of 3.99$^{\circ}$ indicating that regardless of input size, GazeTR achieves ~4.00$^{\circ}$. Interestingly, ``SwinIR-2x" achieved a POG of 3.90$^{\circ}$ and demonstrated a 2.3\% improvement over the $448 \times 448$ baseline, showing greater relative performance when the super-resolved image is not compressed. Furthermore, ``SwinIR-2x" achieved a 1\% improvement in performance over the original $448 \times 448$ high-resolution data (POG of 3.94$^{\circ}$). The results suggest that SR can demonstrate better performance even with lower-resolution data. These findings also suggest that denoising and sharpening images play a larger role than merely increasing the image size. Finally, SR preprocessing achieves state-of-the-art results and beats GazeTR \cite{cheng2021gaze} by 2.3\%.

\begin{table}[t]
\centering
\begin{tabular}{l|cccc}
\hline
Method                                & Input & Gaze        & POG                    \\ \hline
Full-Face \cite{zhang2017s}           & 224& 224         & 4.93$^{\circ}$          \\
RT-Gene  \cite{fischer2018rt}         & 224& 224         & 4.66$^{\circ}$          \\
Dilated-Net \cite{chen2019appearance} & 224& 224         & 4.42$^{\circ}$          \\
Gaze360  \cite{kellnhofer2019gaze360} & 224& 224         & 4.06$^{\circ}$          \\
Gaze360  \cite{kellnhofer2019gaze360} & 224& 224         & 4.06$^{\circ}$          \\
CA-Net     \cite{cheng2020coarse}     & 224& 224         & 4.27$^{\circ}$          \\
Mnist   \cite{zhang2015appearance}    & 224& 224         & 6.39$^{\circ}$          \\
GazeNet     \cite{zhang2017mpiigaze}  & 224 & 224         & 5.76$^{\circ}$          \\
GazeTR-Pure \cite{cheng2021gaze}      & 224 & 224         & 4.74$^{\circ}$          \\
GazeTR-Hybrid \cite{cheng2021gaze}    & 224 & 224         & 4.00$^{\circ}$          \\
Proposed \footnotesize SwinIR-2x Downsampled                & 224 & 224       & \textbf{3.94$^{\circ}$} \\ 
Proposed \footnotesize SwinIR-2x                             & 224 & 448        & \textbf{3.90$^{\circ}$} \\ \hline
\end{tabular}
\caption[GazeTR]{Comparison with prior works. The gaze column denotes the size of the input into the gaze model. }
\label{exp2_tab}
\end{table}

\subsection{Low-Resolution and Degraded Images}
To further elaborate on the findings from Table \ref{exp2_tab}, the performance of SR on low-resolution and degraded images was examined. Additionally, while reproducing results of GazeTR \cite{cheng2021gaze}, it was discovered that a significant portion of its performance was attributed to it's pretraining on the ETH-XGaze dataset \cite{zhang2020eth}. Thus, the effect of pretraining the gaze model was also investigated.

\begin{figure}[h]
\centering
\includegraphics[width=0.45\textwidth]{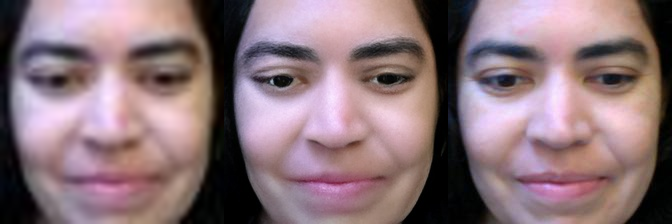}
\caption[Looking Sharp]{ The left is an image downsampled to size 56, the middle is the previous image super-resolved to 224, and the original 224 image.}
\label{56_224}
\end{figure}

To simulate a low-resolution setting, images were downsamped to $112 \times 112$ and also $56 \times 56$ (8x smaller than the original dataset) allowing for a comparison between SR and interpolation upscaling. The results of these experiments are shown in Table \ref{exp3_tab}. To simulate real-world degradations such as motion blur, pixelation and compression, BSR-GAN's complex degradation method \cite{zhang2021designing} was used and again compare SR to interpolation. Furthermore, these three experiments were repeated for GazeTR without pretraining since pretraining on a massive dataset is unrealistic and might skew performance. The pretrained experiments can be found in the second section of Table \ref{exp3_tab} and Table \ref{exp4_tab}. While it is not surprising that pretrained models outperformed non-pretrained, SR exhibited improved performance relative to the interpolation baseline, irrespective of pretraining, initial image resolution or the addition of complex degradations. The experiments carried out emphasize the application of SR as a preprocessing technique, particularly in scenarios where facial images are low-resolution (Figure \ref{56_224}) or have been affected by real-world degradations such as motion blur, pixelation, and compression (Figure \ref{SwinIR}).

\begin{table}[h]
\centering
\begin{tabular}{ccccc}
\hline
Pretraining                      & Interpolation Type & Input  & Gaze & POG  \\ \hline
                                 & Interpolation      & 56  & 224  & 4.81$^{\circ}$          \\
                                 & SwinIR-4x          & 56  & 224  & \textbf{4.76$^{\circ}$} \\
                                 & Interpolation      & 112 & 448  & 4.53$^{\circ}$          \\ 
\multirow{-4}{*}{No}             & SwinIR-4x          & 112 & 448  & \textbf{4.48$^{\circ}$} \\ \hline
                                 & Interpolation      & 56  & 224  & 4.31$^{\circ}$           \\
                                 & SwinIR-4x          & 56  & 224  & \textbf{4.22$^{\circ}$}  \\
                                 & Interpolation      & 112 & 448  & 4.24$^{\circ}$          \\
\multirow{-4}{*}{Yes}            & SwinIR-4x          & 112 & 448  & \textbf{4.21$^{\circ}$} \\ \hline
\end{tabular}
\caption[GazeTR and Low-Resolution Data]{Results on low-resolution images. GazeTR is used as the gaze model.}
\label{exp3_tab}
\end{table}

\begin{table}[h]
\centering
\begin{tabular}{ccccc}
\hline
Pretraining                   & Interpolation Type & Input  & Gaze & POG                       \\ \hline
                              & Interpolation      & 112 & 448  & 5.40$^{\circ}$              \\
\multirow{-2}{*}{No}          & SwinIR-4x          & 112 & 448  & \textbf{5.33$^{\circ}$}    \\ \hline
                              & Interpolation      & 112 & 448  & 5.37$^{\circ}$        \\
\multirow{-2}{*}{Yes}         & SwinIR-4x          & 112 & 448  & \textbf{5.20$^{\circ}$}    \\ \hline
\end{tabular}
\caption[GazeTR and Degraded Data]{Results on degraded images. GazeTR is used as the gaze model.}
\label{exp4_tab}
\end{table}

\section{Training with Less Labelled Data: a Self-supervised Approach based on SR}
Research has previously shown that self-supervised learning can be an effective method for pretraining a backbone network that can be used for various downstream tasks such as object detection, instance segmentation, and semantic segmentation \cite{pmlr-v119-chen20j}. With this in mind, we examine SR for gaze prediction through the lens of self-supervision. The SR tasks doesn't require any explicit labelling of the data. In particular, we propose to obtain a backbone network trained for SR on a large unlabelled dataset and subsequently use it with a relatively simple head trained on small labelled gaze dataset. The result is a sample efficient gaze model that can achieve competitive results but with much less labelled training data. This in turn opens up possibility of building gaze models for scenarios where obtaining a large labelled gaze dataset is challenging, for example gaze tracking in infants \cite{franchak2016free} or older adults \cite{chapman2006evidence}. Another area that may benefit from this is gaze tracking in animals \cite{wiltschko2015mapping, ogura2020dogs, guo2003monkeys}.

\begin{figure}[t]
\centering
\includegraphics[width=0.5\textwidth]{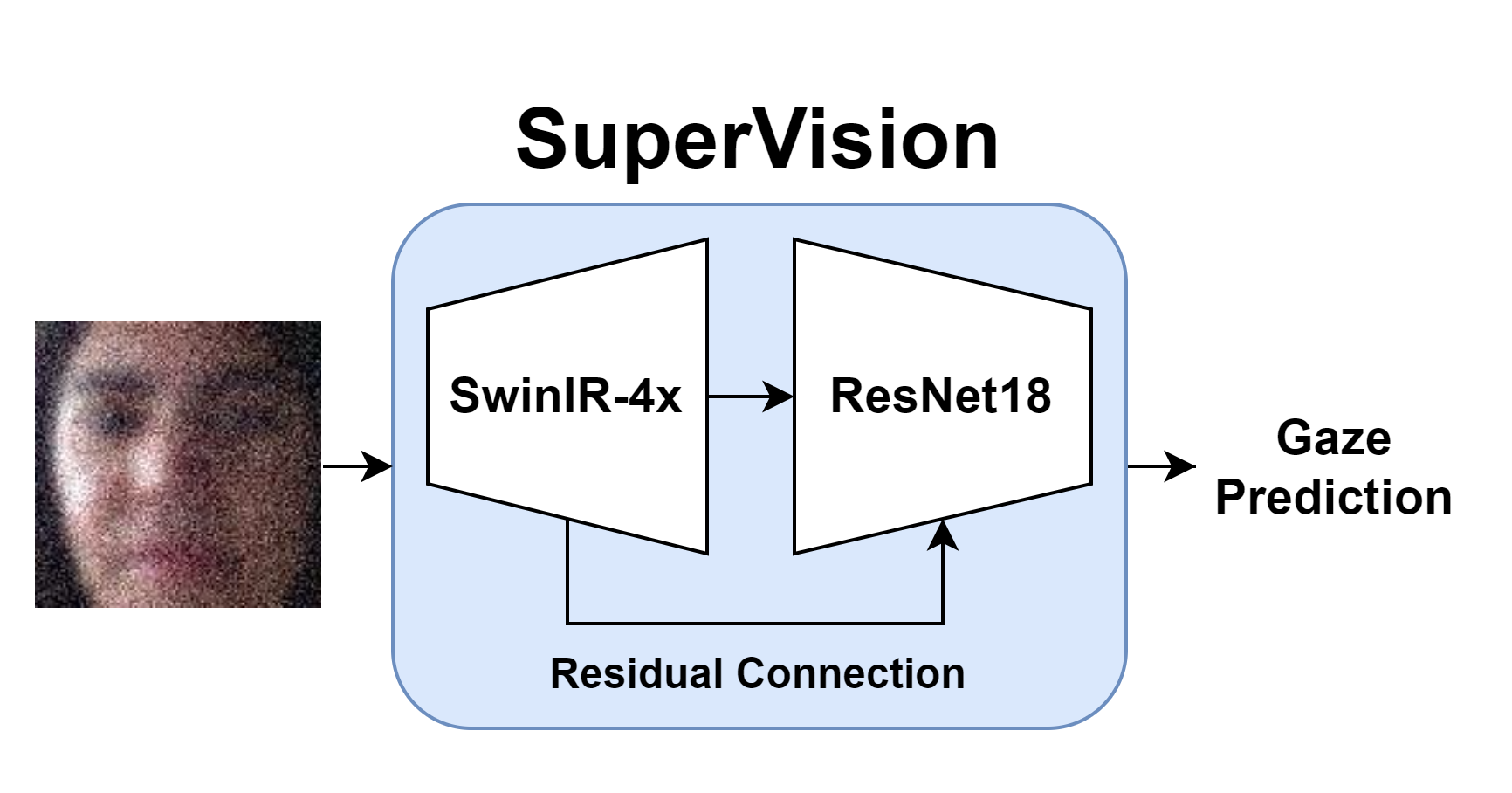}
\caption[SuperVision Architecture]{SuperVision architecture. }
\label{superVision}
\end{figure}

The intermediate representations extracted during the SR process might capture useful patterns and features for gaze prediction. Therefore, we propose a new architecture called ``SuperVision'' by adding residual connections from earlier layers in the SR module to later layers in the Gaze model. Concretely, the SwinIR-4x model with a ResNet18 are concatenated to produce an end-to-end model for appearance-based gaze tracking. The SR-Gaze model may lose auxiliary information during the compression of the SR output to an image. This bottleneck, coupled with the model's depth, justifies utilizing a residual connection from earlier layers in the SR module to later layers in the Gaze model.

\begin{table}[b]
\centering
\begin{tabular}{c|ccc}
\hline
Interpolation Type               & \multicolumn{3}{c}{POG}                                               \\ \hline                
Interpolation            & 6.26$^{\circ}$            & 6.06$^{\circ}$           & 6.04$^{\circ}$          \\
SwinIR-4x  Downsampled              & 6.20$^{\circ}$            & 6.01$^{\circ}$           & 5.91$^{\circ}$          \\
SuperVision                      & \textbf{6.17$^{\circ}$}   & \textbf{5.90$^{\circ}$}  & \textbf{4.54$^{\circ}$} \\ \hline
\multicolumn{1}{c|}{Data Used} & \multicolumn{1}{c}{5\%}   & \multicolumn{1}{c}{10\%} & 20\%   \\ \hline
\multicolumn{4}{c}{\footnotesize *For reference, POG of GazeTR with 100\% of the data is 5.37$^{\circ}$ }
\end{tabular}
\caption[SuperVision]{ Training with a small portion of the training data. A simple ResNet18 is used as the gaze model. The proposed SuperVision method uses 5x less labeled data and yet outperforms the state-of-the-art method of GazeTR which uses 100\% of training data.}
\label{exp5_tab}
\end{table}

 To evaluate the effectiveness of the proposed method, we run experiments and compare it with SR preprocessing and an interpolation baseline. For this experiment low-resolution images of size 112 were used so that the SwinIR's 4x upsampling outputs a reasonable $448 \times 448$ size. These images are then downsampled as ResNet18 accepts images of size 224. We run experiments with 5\%, 10\% and 20\% of the labelled data. As expected, results in Table \ref{exp5_tab} show a trend of improvement as more labelled training data are used. Particularly noteworthy is the rapid acceleration of relative improvement over the interpolation baseline, especially for SuperVision, which achieved a POG of 4.54 and demonstrated a 33\% improvement at 20\% training data. The SuperVision model was also evaluated without a residual connection, and the results indicated a 4.6\% improvement with the inclusion of a residual connection. Moreover, we can directly compare SuperVision's results to those of Table \ref{exp4_tab} since both models use the same data of 112 resolution with complex degradations. Despite using only 20\% of the data, SuperVision displayed a 15\% improvement over the state-of-the-art model of GazeTR which is pretrained on the full ETH-XGaze dataset and also uses 100\% of the labeled data in MPIIFaceGaze dataset. These experiments demonstrate SR can be used to achieve a high performance with only a fraction of labelled data.

\section{Limitations}
One of the primary challenges we encountered during the training phase was hardware limitations. Due to these constraints, we had to rely on smaller SR models such as SwinIR-2x instead of SwinIR-4x for several experiments. The reason being that scaling up from $224 \times 224$ by 4x would not have been feasible on our Tesla V100 GPUs.
%Additionally, training our end-to-end model SuperVision on all the samples became impractical due batch size limitations leading to significant training time.

\section{Conclusion}
We explored the potential of SR as a preprocessing step for appearance-based gaze prediction. The study demonstrated that not all SR models are effective in preserving gaze direction. However, the proposed two-step framework based on SR achieved state-of-the-art results on the MPIIFaceGaze dataset. The method was also evaluated on low-resolution and degraded images which confirmed the effectiveness of SR. Moreover, the newly introduced SuperVision architecture, which combines SR and ResNet with skip connections, showed impressive results using only 20\% of the labeled data, outperforming the state-of-the-art GazeTR method which uses 100\% of the data by 15\%. These findings demonstrate the potential of the proposed SuperVision method as a more efficient and effective approach to appearance-based gaze prediction.

%To support the assertion that SR can serve as an efficient preprocessing tool, additional work might involve applying SR preprocessing to various other computer vision tasks.

%%%%%%%%% REFERENCES
{\small
\bibliographystyle{ieee_fullname}
\bibliography{egbib}
}

\end{document}